\documentclass[journal ]{new-aiaa}
\usepackage[utf8]{inputenc}
\usepackage{textcomp}

\usepackage{graphicx}
\usepackage{amsmath}
\usepackage[version=4]{mhchem}
\usepackage{siunitx}
\usepackage{longtable,tabularx}
\usepackage{subcaption}
\usepackage{minted}
\usepackage[margin=1in]{geometry}
\usepackage{tikz}
\usepackage{comment}
\usepackage{cleveref}
\usepackage{tikz}
\usetikzlibrary{arrows.meta, calc}

\usetikzlibrary{shapes,arrows,positioning}

\title{Optimal Reward Shaping: Autonomous Car Parking Case Study}

\author{Emre Özkaya\footnote{Senior researcher.} and Nicolas R. Gauger\footnote{Professor}}
\affil{Chair for Scientific Computing, RPTU University of Kaiserslautern-Landau \\ Paul-Ehrlich-Strasse 34, 67663 Kaiserslautern}

\begin{document}

\maketitle

\begin{abstract}
Designing effective reward functions for model-free reinforcement learning under non-holonomic constraints remains a persistent challenge, often resulting in severe local minima such as policy paralysis or over-conservative hazard avoidance. In this work, we present a parameterized reward shaping framework featuring coverage-gated alignment feedback, drive-direction switch regularization, and an aligned episode termination mechanism evaluated on an autonomous parallel parking task. Crucially, we show that environmental reward parameters and algorithmic hyperparameters are deeply co-dependent, requiring joint meta-optimization to achieve stable convergence. By employing surrogate-based Bayesian optimization, our co-optimized Deep Q-Network (DQN) agent resolves characteristic control failure modes, significantly outperforming uncalibrated baselines across both success rate and trajectory smoothness.
\end{abstract}

\section{Introduction}
In the field of reinforcement learning (RL), designing an effective reward function remains one of the most critical yet challenging obstacles to deploying successful policies. While model-free reinforcement learning has demonstrated remarkable success across varied application domains, its performance is heavily bound to the quality of the feedback signal provided to the agent. In complex physical environments, standard sparse reward structures often lead to intractable exploration times or catastrophic gradient stagnation. Conversely, poorly constructed dense rewards frequently introduce unintended local minima, causing the agent to optimize for immediate rewards rather than long-term task completion. This fundamental challenge makes the study of optimal reward shaping a cornerstone of reliable, sample-efficient RL.

To evaluate these structural reward challenges under severe physical constraints, autonomous vehicle parking serves as an ideal case study \cite{song2020data}. Precision docking maneuvers---such as parallel and reverse parking---require a vehicle to operate within highly confined spatial corridors. This task is intensely complicated by non-holonomic Ackermann steering kinematics, meaning the vehicle cannot translate sideways and must execute complex, multi-stage trajectory corrections \cite{berta2024development}. Traditional dense reward formulations, which naively aggregate linear combinations of spatial distance errors, orientation misalignments, and negative collision penalties, fail catastrophically in this domain due to two critical optimization anomalies \cite{tian2026hierarchical}. 

First, a static linear combination introduces a severe objective mismatch: an agent far away from its destination is penalized for its heading angle prematurely, leading to erratic local minima such as spinning in circles near or far from the target zone \cite{tian2026hierarchical}. Second, the initialization of the collision penalty creates a highly fragile boundary between exploration and risk aversion. If the penalty for crashing is set too low, the agent continuously sacrifices structural integrity for minor progress gains; conversely, if the collision penalty is excessively harsh, the value function network rapidly succumbs to a "cautious coward" policy \cite{sutton2018reinforcement}. Under this failure mode, the agent learns to safely avoid collisions by freezing in place or executing defensive maneuvers entirely outside the parking zone, optimizing to minimize penalty streams rather than attempting to park.

This paper addresses the reward design crisis by presenting a comprehensive framework for optimal reward shaping, utilizing a custom 2D Ackermann parking environment as a proving ground. We discard static reward mixtures in favor of a parameterized reward function. Crucially, we deliberately avoid the introduction of external behavioral heuristics or algorithmic constraints, such as action masking or kinematic pruning \cite{huang2020closer,abel2015goal}. While such techniques would inherently inflate empirical convergence metrics by artificially narrowing the exploration space, doing so would obscure the core scientific objective of this study. Because our primary purpose is to expose, isolate, and systematically address the fundamental vulnerabilities of reward shaping under non-holonomic constraints, the parking environment is treated as a clean, unassisted baseline. Furthermore, we mitigate value function mismatches by mathematically binding the environment's early termination success conditions directly to the comprehensive final evaluation score, ensuring that step-by-step rewards remain strictly aligned with ultimate completion criteria.

Importantly, our case study demonstrates that optimizing the reward framework cannot be achieved through isolated intuition or default algorithmic configurations. The interplay between non-linear potential field scaling and deep network exploration schedules creates a highly sensitive, non-convex optimization landscape. We show that single-run training sessions utilizing off-the-shelf model hyperparameters consistently fail to converge. Consequently, automated Hyperparameter Optimization (HPO) is integrated into our architecture not merely as an optional step for minor performance tuning, but as a foundational prerequisite to map the co-dependent landscape of reward constants and policy update variables.

\subsection{Contributions}
The primary contributions of this work are threefold:
\begin{enumerate}
    \item \textbf{Parameterized Reward Function Formulation:} We establish a parameterized reward structure incorporating spatial distance penalties, coverage-gated alignment feedback, and drive-direction switch regularization, enabling dynamic shifting of objective priorities from macro-navigation to high-precision angular docking while suppressing high-frequency control chatter.
    \item \textbf{Unified Termination Foundations:} We introduce a mathematically aligned episode cutoff mechanism that eliminates the common reinforcement learning ``dead zone'' problem by directly tying early success termination to the continuous multi-criteria evaluation metric.
    \item \textbf{Co-Optimization Landscape Analysis:} Through a systematic hyperparameter optimization case study, we map and analyze the strict co-dependencies between environmental reward parameters and deep Q-network learning variables, proving that joint optimization is a mandatory structural requirement for non-holonomic control tasks.
\end{enumerate}

\section{Problem Formulation and System Model}
The primary objective of the autonomous parking test case is to train an ego-vehicle to safely navigate from an arbitrary road spawn coordinate and execute a precise, collision-free reverse docking maneuver into a designated parking area. Success is defined by achieving a strict spatial match with the target center point while aligning the vehicle's heading parallel to the curb, requiring the agent to coordinate tight low-speed steering adjustments. To evaluate the efficiency of the hyperparameter meta-optimization layer in overcoming these physical limits, this trajectory task must be framed within the context of model-free reinforcement learning (RL). We model the vehicle's control loop as a sequential decision process governed by a Markov Decision Process (MDP), defined by the tuple $\mathcal{M} = \langle \mathcal{S}, \mathcal{A}, \mathcal{P}, \mathcal{R}, \gamma \rangle$. Here, $\mathcal{S}$ represents the continuous state space, $\mathcal{A}$ defines the discrete action space, $\mathcal{P}: \mathcal{S} \times \mathcal{A} \to \Delta(\mathcal{S})$ denotes the environmental state transition dynamics mapping state-action pairs to a probability distribution over subsequent states, $\mathcal{R}: \mathcal{S} \times \mathcal{A} \to \mathbb{R}$ is the dynamically parameterized shaped reward function, and $\gamma \in [0, 1)$ is the temporal discount factor.

\subsection{Environmental Topology and Spatial Constraints}
The simulation environment is hosted within a bounded $800 \times 800$, pixel arena designed to replicate a constrained urban parallel parking scenario, as illustrated in Figure~\ref{fig:parking_arena}. The topology is explicitly divided into functional lanes and structural hazards:

\begin{figure}[H]
    \centering
    \includegraphics[width=0.55\textwidth]{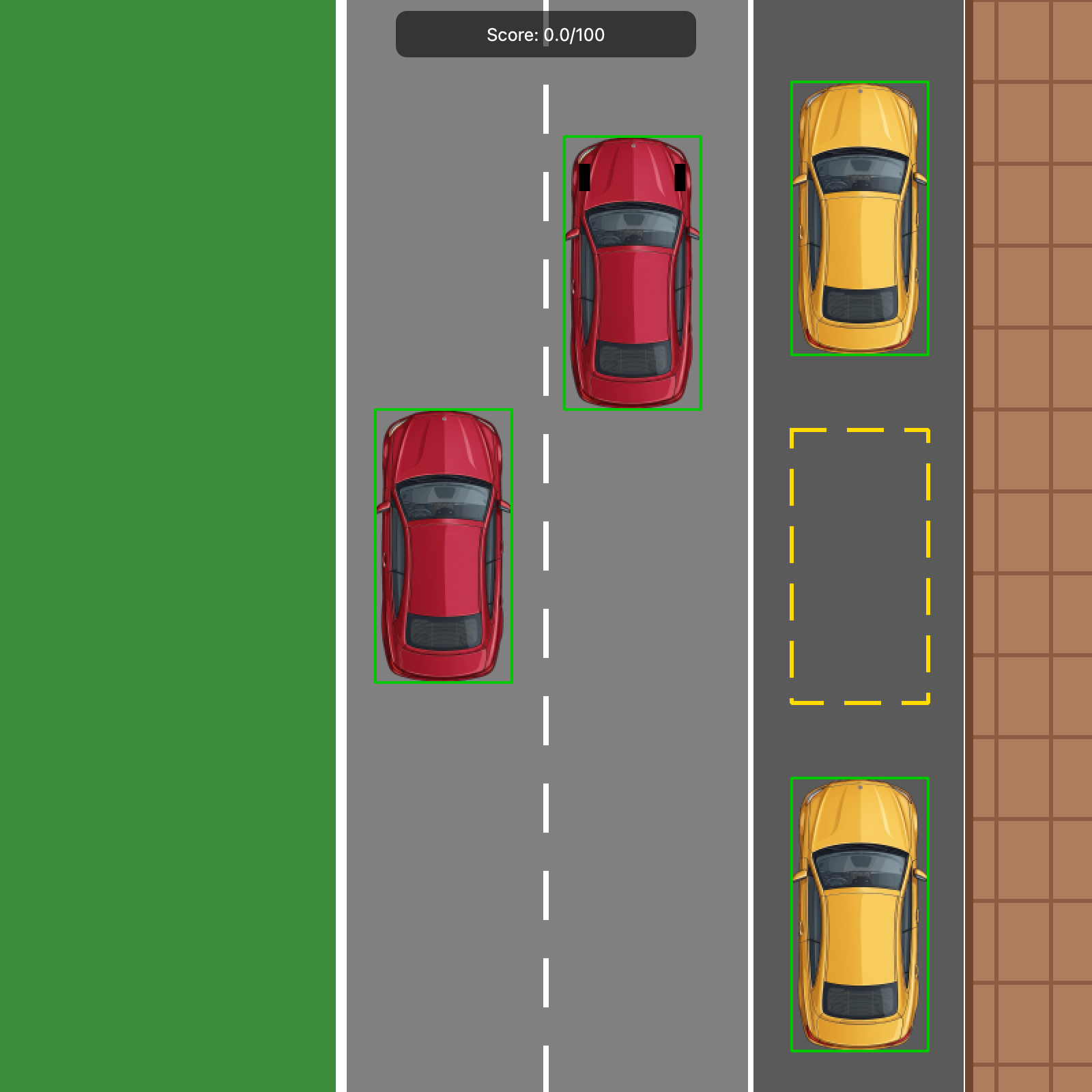}
    \caption{The complete simulation game board configuration layout. The diagram illustrates the vehicle's initial roadway coordinate span, the impassable road boundary, the lateral right pavement curb line, and the target parking stall centered perfectly between the two static obstacle vehicles.}
    \label{fig:parking_arena}
\end{figure}

\begin{itemize}
    \item \textbf{Active Roadways and Boundaries:} The driving corridor is bounded on the left by an impassable road margin at $x = 250\,\text{px}$ (green zone) and on the right by a strict pavement curb line situated at $x = 710\,\text{px}$ (terracotta-tinted zone).
    \item \textbf{Dynamic Traffic Hazard:} The left lane includes an active, moving traffic vehicle centered at a lateral coordinate of $x = 325.0\,\text{px}$. This obstacle moves dynamically down the roadway at a constant velocity of $80.0\,\text{px/s}$, automatically resetting its longitudinal position upon boundary exit to generate continuous, non-stationary environmental disruptions that the ego-agent must actively evade.
    \item \textbf{Static Obstacles (Yellow Parked Vehicles):} The parking lane features two invariant, stationary vehicles that bound the target stall. Each obstacle has a standardized width of $100\,\text{px}$ and a height of $200\,\text{px}$. These cars are anchored along the curb at $x = 630.0\,\text{px}$, centered at longitudinal positions $y_1 = 160.0\,\text{px}$ and $y_2 = 670.0\,\text{px}$.
    \item \textbf{Target Parking Spot (Yellow Dashed Zone):} The ideal docking zone lies within the pocket created by the parked vehicles. The absolute target center is centered at $(x_{\text{target}}, y_{\text{target}}) = (630.0\,\text{px}, 415.0\,\text{px})$ with a target heading directed straight down the lane ($\theta_{\text{target}} = -\frac{\pi}{2}$).
\end{itemize}

\subsection{Vehicle Kinematics and Non-Holonomic Constraints}
\begin{figure}[htbp]
\centering
\begin{tikzpicture}[scale=0.04, >=Stealth]

    \draw[fill=black!12, draw=black!70, thick] (-50, -40) rectangle (50, 160);

    \draw[line width=1.5pt, blue!60!black] (-50, 0) -- (50, 0);     
    \draw[line width=1.5pt, blue!100!black] (-50, 120) -- (50, 120); 
    \draw[line width=1.5pt, blue!60!black] (0, 0) -- (0, 120);       

    \draw[fill=blue!80!black, rounded corners=2pt] (-58, -18) rectangle (-42, 18);
    \draw[fill=blue!80!black, rounded corners=2pt] (42, -18) rectangle (58, 18);

    \begin{scope}[shift={(50, 120)}, rotate=30]
        \draw[fill=blue!80!black, rounded corners=2pt] (-8, -18) rectangle (8, 18);
    \end{scope}
    \begin{scope}[shift={(-50, 120)}, rotate=30]
        \draw[fill=blue!80!black, rounded corners=2pt] (-8, -18) rectangle (8, 18);
    \end{scope}

    \filldraw[blue!90!black] (0,0) circle (3.5) node[below right, black, xshift=2pt] {};
    \filldraw[blue!90!black] (0,120) circle (3.5);

    \draw[|<->|] (-75, -40) -- (-75, 160) node[midway, fill=white, rotate=90] {200 px};
    \draw[|<->|] (-50, -65) -- (50, -65) node[midway, fill=white] {100 px};

    \draw[dashed, black!40] (0, 120) -- (0, 170);
    \draw[dashed, black!40] (0, 120) -- ++(120:50); 
    \draw[->, orange, thick] (0, 155) arc (90:120:35) node[midway, above left] {$\delta$ };

\end{tikzpicture}
\caption{Kinematic bicycle model.}
\label{fig:kinematics_bicycle}
\end{figure}
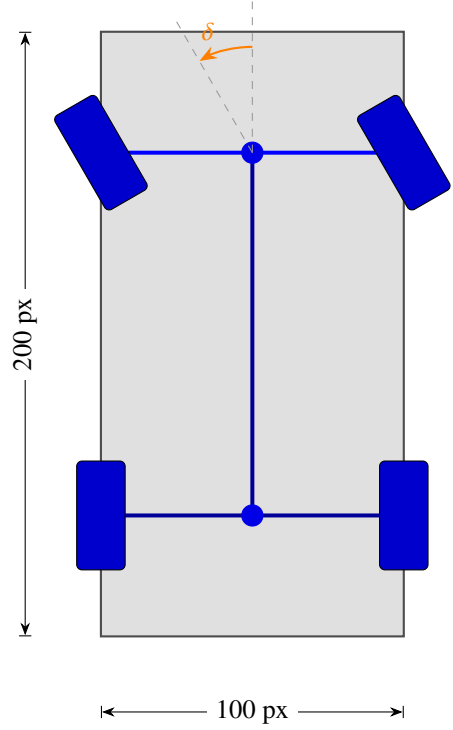

The ego-vehicle is modeled as a rigid rectangular bounding body ($100 \times 200\,\text{px}$) operating under non-holonomic kinematics, as illustrated in Figure~\ref{fig:kinematics_bicycle}. The continuous system state vector is defined as $q = [x, y, \theta, \delta]^T$, where $(x, y)$ represents the planar Cartesian coordinate of the rear-axle center, $\theta$ denotes the global heading orientation relative to the horizontal axis, and $\delta$ tracks the instantaneous steering articulation angle. 

The movement of the vehicle is constrained by the kinematic bicycle model approximation. Due to the non-holonomic rolling constraints, the vehicle possesses zero lateral velocity in its local reference frame, meaning it cannot translate sideways without longitudinal motion. Under an active longitudinal velocity $v$ and steering angle $\delta$, the continuous equations of motion are governed by:
\begin{align}
\dot{x} &= v \cos(\theta) \\
\dot{y} &= v \sin(\theta) \\
\dot{\theta} &= \frac{v}{L} \tan(\delta)
\end{align}
where $L = 60.0\,\text{px}$ denotes the vehicle wheelbase length separating the front and rear axles. 

In our simulated reinforcement learning environment, these continuous dynamics are discretized over a fixed execution timestep $\Delta t = 0.1\,\text{s}$ using Euler integration:
\begin{align}
x_{t+1} &= x_t + v_t \cos(\theta_t) \cdot \Delta t \\
y_{t+1} &= y_t + v_t \sin(\theta_t) \cdot \Delta t \\
\theta_{t+1} &= \theta_t + \frac{v_t}{L} \tan(\delta_t) \cdot \Delta t
\end{align}
In accordance with physical hardware limitations, the internal steering angle is hard-clamped such that $\delta \in [-\delta_{\max}, \delta_{\max}]$, where the steering limit is defined as $\delta_{\max} = \frac{\pi}{6}$ radians ($30^\circ$).

\subsection{Action Space and Discrete Joint Control}
To facilitate discrete action selection while supporting underlying continuous steering physics, the action space $\mathcal{A}$ is formulated as a factorized discrete set of joint maneuvers:
\begin{equation}
\mathcal{A} = \{0, 1, 2, \dots, 8\}, \quad |\mathcal{A}| = 9
\end{equation}
Operating at a fixed physics simulation increment of $\Delta t = 0.016\,\text{s}$ per environment step, each action maps uniquely to a combined longitudinal velocity command $v$ and an incremental lateral steering direction $\text{dir}_{\text{steer}} \in \{-1, 0, 1\}$ calculated via $\text{dir}_{\text{steer}} = (\text{action} \pmod 3) - 1$:

\begin{enumerate}
    \item \textbf{Actions 0--2 (Forward Propulsion):} Commands a positive constant velocity $v = 100.0\,\text{px/s}$. Action 0 steers maximum left ($\text{dir}_{\text{steer}} = -1$), Action 1 commands neutral straight tracking ($\text{dir}_{\text{steer}} = 0$), and Action 2 steers maximum right ($\text{dir}_{\text{steer}} = +1$).
    \item \textbf{Actions 3--5 (Reverse Propulsion):} Commands a backward constant velocity $v = -100.0\,\text{px/s}$. Steering transitions analogously from left (Action 3), straight (Action 4), to right (Action 5). This mode is essential for performing the classic backward parallel parking entry maneuver.
    \item \textbf{Actions 6--8 (Stationary / Active Braking):} Halts longitudinal progression ($v = 0.0\,\text{px/s}$). Actions 6, 7, and 8 isolate wheel-turning dynamics while stationary, adjusting the steering angle left, neutral, or right, respectively.
\end{enumerate}

The steering angle update is incrementally adjusted by a steering velocity magnitude of $45^\circ/\text{s}$ ($\frac{\pi}{4}\,\text{rad/s}$) and structurally clamped to avoid non-physical joint over-rotation:
\begin{equation}
\delta_{t+1} = \max\left(-\frac{\pi}{6}, \, \min\left(+\frac{\pi}{6}, \, \delta_t + \text{dir}_{\text{steer}} \times \frac{\pi}{4} \times \Delta t\right)\right)
\end{equation}
This formulation forces the learning agent to strategically coordinate its speed selection and wheel placement to avoid hard collisions within the tightly bounded parking lane.

\section{Observation Space and State Vector Architecture}

To satisfy the Markov property, the observation vector must provide the learning agent with comprehensive spatial awareness, target-relative tracking tracking coordinates, and directional proximity feedback. We construct a 21-dimensional continuous state vector, $o_t \in \mathbb{R}^{21}$. To ensure stable neural network gradient updates and prevent feature dominance issues, all continuous variables are normalized to a strict $[-1.0, 1.0]$ or $[0.0, 1.0]$ operational range. 

The structural layout of the state vector is factorized into two primary functional components: core kinematic features ($9D$) and a sensory proximity LIDAR array ($12D$). The complete observation vector is defined as:

\begin{equation}
o_t = \Big[ \underbrace{\bar{x}, \, \bar{y}, \, \sin(\theta), \, \cos(\theta), \, \Delta \bar{x}, \, \Delta \bar{y}, \, \bar{x}_{\text{curb}}, \, \bar{\delta}, \, \bar{v}}_{\text{Core Kinematics } (9D)}, \,\, \underbrace{L_1, \, L_2, \, \dots, \, L_{12}}_{\text{Proximity LIDAR } (12D)} \Big]^T
\end{equation}

\subsection{Core Kinematical and Relational Features}
The first nine dimensions capture the immediate physical state of the ego-vehicle and its explicit spatial relationship to both the road limits and the final docking target:

\begin{enumerate}
    \item \textbf{Absolute Vehicle Position ($\bar{x}, \bar{y}$):} Represents the planar Cartesian center coordinates of the vehicle, scaled linearly against the maximum dimensions of the simulation arena ($800.0$ pixels):
    \begin{equation}
    \bar{x} = \frac{x_{\text{car}}}{800.0}, \quad \bar{y} = \frac{y_{\text{car}}}{800.0}
    \end{equation}
    
    \item \textbf{Decomposed Orientation ($\sin(\theta), \cos(\theta)$):} Instead of passing the raw heading angle $\theta$ directly—which introduces a sharp mathematical step discontinuity at the $\pm\pi$ boundary wrap-around—we decompose the orientation into its trigonometric components. This ensures a smooth, continuous topology across a full $360^\circ$ rotation matrix.
    
    \item \textbf{Relative Distance to Target Center ($\Delta \bar{x}, \Delta \bar{y}$):} Explicit tracking errors computed relative to the invariant target destination center ($(x_{\text{target}}, y_{\text{target}}) = (630.0, 415.0)$), normalized against the arena scale:
    \begin{equation}
    \Delta \bar{x} = \frac{x_{\text{target}} - x_{\text{car}}}{800.0}, \quad \Delta \bar{y} = \frac{y_{\text{target}} - y_{\text{car}}}{800.0}
    \end{equation}
    
    \item \textbf{Curb Proximity Indicator ($\bar{x}_{\text{curb}}$):} Explicitly tracks the remaining lateral clearance between the vehicle center and the right pavement curb line ($710.0$ pixels):
    \begin{equation}
    \bar{x}_{\text{curb}} = \frac{710.0 - x_{\text{car}}}{800.0}
    \end{equation}
    
    \item \textbf{Normalized Actuator States ($\bar{\delta}, \bar{v}$):} Represents the instantaneous configuration of the vehicle's mechanics from the preceding environment step. The internal steering angle is scaled against its maximum mechanical limit ($30^\circ \approx 0.523$\,rad), and the velocity is normalized against the maximum propulsion velocity ($100.0$\,pixels/s):
    \begin{equation}
    \bar{\delta} = \frac{\delta_t}{\delta_{\max}}, \quad \bar{v} = \frac{v_t}{100.0}
    \end{equation}
\end{enumerate}

\subsection{Proximity Sensing via Multi-Ray LIDAR Array}
\begin{figure}[H]
    \centering
    \includegraphics[width=0.55\textwidth]{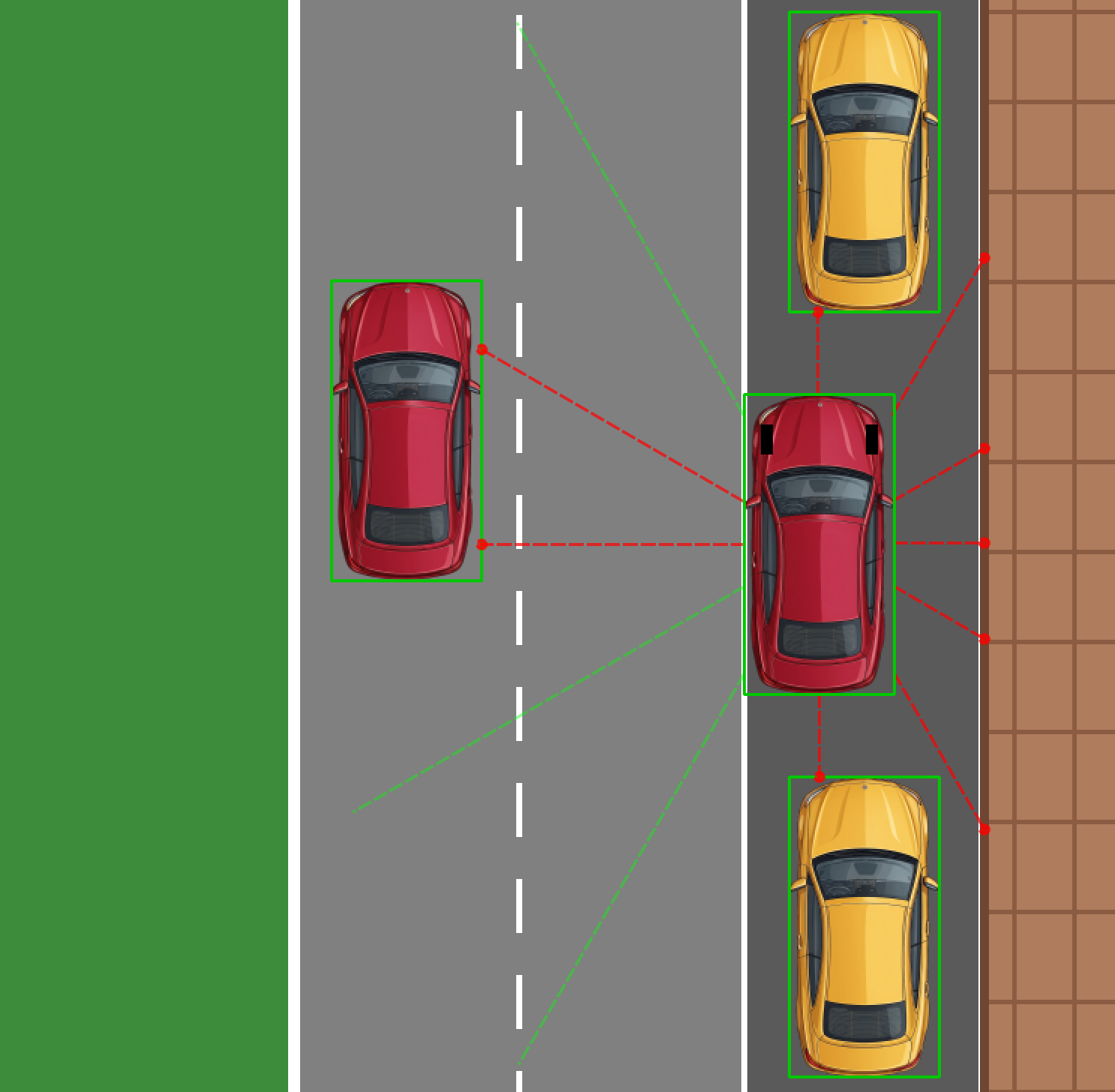}
    \caption{The omnidirectionally projected 12-ray planar LIDAR array. The lines illustrate the spatial outward projection from the vehicle's rectangular bounding frame edges, tracing intersection points with adjacent parked cars and roadway boundaries.}
    \label{fig:lidar_rays}
\end{figure}

To navigate successfully around static and dynamic obstacles without hard-coding the coordinates of nearby vehicles as explicit oracle features, the state vector incorporates a 12-ray planar LIDAR array (Dimensions 10–21). The physical layout and intersection behavior of these rays are illustrated in Figure~\ref{fig:lidar_rays}.

The rays project outward omnidirectionally from the physical rectangular bounding box edges of the vehicle at uniform angular increments of $30^\circ$ relative to the current vehicle heading $\theta$. Each individual ray computes the closest intersection distance to a boundary hazard (such as the road margin, pavement curb, adjacent parked cars, or active traffic) using a ray-AABB (Axis-Aligned Bounding Box) calculation step. The raw intersection distance is capped at a maximum range horizon of $R_{\max} = 300.0$ pixels and scaled linearly:
\begin{equation}
L_i = \frac{\text{Distance to obstacle}_i}{R_{\max}}, \quad L_i \in [0.0, 1.0]
\end{equation}
A normalized reading of $1.0$ indicates an entirely clear operational path along that ray's angular vector, while a reading of $0.0$ indicates an immediate obstacle impact occurring directly at the physical edge of the vehicle frame. This dense geometric representation provides model-free architectures with an expressive spatial footprint to map non-holonomic, collision-free paths.

\section{Parameterized Dynamic Reward Function}

To overcome the severe sample inefficiency associated with sparse terminal objectives, the environment implements a hybrid composite reward function $\mathcal{R}(s_t, a_t)$. This framework explicitly integrates continuous, dense feedback channels and temporal action regularization penalties with event-driven sparse boundary points. By providing immediate step-by-step guidance alongside discrete programmatic incentives at every simulation increment $\Delta t$, the reward structure generates a smooth optimization gradient that guides the non-holonomic ego-agent from its initial roadway spawn coordinates toward a precise, stable park.

The global step-by-step reward calculation combines a constant time penalty, continuous spatial tracking surrogates, drive-direction switch regularization, and sparse, discrete terminal conditions. It is evaluated at each environment step as follows:
\begin{equation}
\mathcal{R}(s_t, a_t) = R_{\text{time}} + R_{\text{switch}} + R_{\text{lane}} + R_{\text{align}} + R_{\text{dist}} + R_{\text{terminal}}
\end{equation}

\subsection{Baseline Operational Penalties}
To encourage temporal efficiency and counteract the potential delays introduced by risk-averse behavior, a continuous baseline time penalty is applied at every simulation step. This dense penalty acts as an optimized meta-parameter, exerting steady negative pressure on the agent to prevent the vehicle from remaining stationary, idling, or spinning indefinitely without approaching the parking area:
\begin{equation}
R_{\text{time}} = -c_{\text{baseline}}
\end{equation}
where $c_{\text{baseline}}$ is the baseline time penalty magnitude dynamically assigned by the hyperparameter meta-optimization layer.

\subsection{Drive-Direction Switch Regularization}
In physical vehicle control, rapid and uncoordinated toggling between driving regimes—such as abrupt transitions between forward propulsion, active braking/neutral, and reverse gear engagement—leads to inefficient, jerky trajectories. From a reinforcement learning perspective, model-free agents operating under discrete joint action spaces frequently exhibit high-frequency chatter between drive states ($v_t \in \{-100.0, 0.0, +100.0\}\,\text{px/s}$) to exploit localized step-wise rewards without executing smooth kinematic arcs.

To regularize policy actions and promote smooth motion profiles, we introduce an explicit drive-direction switch penalty $R_{\text{switch}}$. Let $d_t \in \{-1, 0, 1\}$ denote the discrete longitudinal drive state active at simulation step $t$, corresponding to reverse ($d_t = -1$), stationary/braking ($d_t = 0$), and forward ($d_t = 1$) propulsion. The regularization penalty is evaluated at each step $t > 1$ as:
\begin{equation}
R_{\text{switch}} = \begin{cases}
-c_{\text{switch}} & \text{if } d_t \neq d_{t-1}, \\
0.0 & \text{otherwise.}
\end{cases}
\end{equation}
where $c_{\text{switch}}$ represents the penalty magnitude dynamically tuned as a design variable by the outer hyperparameter meta-optimization layer.

Crucially, this penalty is applied to any transition where the longitudinal drive state changes relative to the preceding step ($d_t \neq d_{t-1}$), explicitly penalizing not only full directional reversals ($1 \to -1$ or $-1 \to 1$), but also intermediate transitions to and from the stationary state ($1 \to 0$ or $0 \to -1$). By penalizing high-frequency state switching, $R_{\text{switch}}$ suppresses erratic gear chatter, encouraging the agent to plan sustained, fluid multi-step maneuvers during complex reverse docking.

\subsection{Spatial Lane-Entry Induction}
Before the vehicle can engage in fine-grained micro-alignment, it must successfully transition from the wide, unconstrained roadway corridor into the highly restricted parallel parking lane. In standard model-free reinforcement learning applications, initiating exploration across an open canvas toward a narrow, obstacle-bounded destination introduces severe gradient sparsity; the agent receives little to no contextual feedback until it accidentally enters the target berth or collides with surrounding hazards. To resolve this exploratory roadblock, we implement a spatial lane-entry induction mechanism that serves as a coarse macro-navigation curriculum.

The environment measures the intersection between the vehicle and the parking zone via a real-time area bounding calculation. Let $C_t \in [0.0, 1.0]$ denote the vehicle's coverage ratio at time step $t$, representing the normalized percentage of the car's rigid rectangular body localized inside the designated parking lane boundaries ($x \in [550, 710]\,\text{px}$). When the vehicle crosses this spatial threshold such that $C_t > 0$, the environment activates a dual-component dense progress reward:
\begin{equation}
R_{\text{lane}} = w_{\text{standing}} \cdot C_t + w_{\text{delta}} \cdot (C_t - C_{t-1})
\end{equation}
where $w_{\text{standing}}$ and $w_{\text{delta}}$ are non-zero weight coefficients treated strictly as optimized parameters managed by the hyperparameter meta-optimization pipeline. 

The explicitly designed double-structure of $R_{\text{lane}}$ addresses two distinct control requirements necessary to successfully bridge the macro-to-micro transition:
\begin{enumerate}
    \item \textbf{Continuous Corridor Stabilization ($w_{\text{standing}} \cdot C_t$):} This term operates as an existence bonus scaled directly to spatial footprint overlap. It provides continuous positive reinforcement that incentivizes the agent to remain inside the parking lane boundaries once entry is achieved, preventing the policy from drifting back out onto the open roadway.
    \item \textbf{Directional Progress Gradient ($w_{\text{delta}} \cdot (C_t - C_{t-1})$):} This term evaluates the instantaneous temporal difference in footprint coverage, $\Delta C = C_t - C_{t-1}$. By functioning as a pseudo-velocity bonus along the lateral approach vector, it injects a strong directional gradient that aggressively rewards the vehicle for moving deeper into the parking lane while penalizing any premature retreat.
\end{enumerate}
By parameterizing this double-structure, the downstream meta-optimization layer can dynamically scale $w_{\text{standing}}$ and $w_{\text{delta}}$ relative to the global collision and time penalties. This establishes a well-behaved potential ramp that safely coaxes the non-holonomic vehicle into the tight parking lane before micro-alignment begins.

\subsection{Distance and Coverage-Dependent Potential Fields}
The primary structural hurdle in non-holonomic vehicle docking is balancing coarse spatial pathfinding toward the berth with high-precision angular alignment within tight geometric bounds. In naive dense formulations, penalizing orientation errors when the vehicle is far from its destination introduces uninformative optimization corridors, causing the agent to fall into local minima such as premature immobilization or spinning. Conversely, omitting alignment guidance until final parking results in crooked poses that violate lane boundaries.

To resolve this trade-off the environment structures step-wise potential feedback around two physical metrics: the Euclidean target distance $d_t = \sqrt{(x_t - x_{\text{target}})^2 + (y_t - y_{\text{target}})^2}$ and the geometric lane coverage ratio $C_t \in [0.0, 1.0]$. The coverage $C_t$ measures the proportion of a $5 \times 3$ sampling grid placed across the vehicle footprint that falls within the designated parking lane corridor ($x \in [550, 710]$).

When the vehicle is outside or near-complete within the lane ($C_t < C_{\text{low}}$ or $C_t > C_{\text{high}}$), distance guidance is enforced linearly relative to the workspace dimension ($800.0\text{ px}$):
\begin{equation}
    R_{\text{dist}} = -\left(\frac{d_t}{800.0}\right) \cdot w_{\text{dist}}
\end{equation}
where $w_{\text{dist}}$ scales the global spatial guidance. 

Once the vehicle is mostly inside the parking lane ($C_t > C_{\text{high}}$), the environment conditionally activates an angular alignment penalty based on the normalized heading difference relative to the target angle $\theta_{\text{target}} = -\pi/2$:
\begin{equation}
    R_{\text{align}} = -\left(\frac{|\Delta\theta_t|}{\pi/2}\right) \cdot w_{\text{align}} \quad \text{for } C_t > C_{\text{high}}
\end{equation}
where $\Delta\theta_t = \text{atan2}(\sin(\theta_t - \theta_{\text{target}}), \cos(\theta_t - \theta_{\text{target}})) \in [-\pi, \pi]$ and $w_{\text{align}}$ enforces orientation alignment as the vehicle completes its parking maneuver.

To encourage stable spatial progression inside the lane corridor, dense coverage rewards are awarded at each timestep ($C_t > 0$):
\begin{equation}
    R_{\text{coverage}} = C_t \cdot w_{\text{standing}} + (C_t - C_{t-1}) \cdot w_{\text{delta}}
\end{equation}
where $w_{\text{standing}}$ (\texttt{standing\_reward\_weight}) provides a continuous existence bonus for staying within the berth, and $w_{\text{delta}}$ (\texttt{delta\_reward\_weight}) provides positive gradient feedback for increasing lane insertion. Combined with a constant baseline time step penalty $R_{\text{time}} = -w_{\text{time}}$, this structured potential field creates a smooth, naturally gated curriculum: broad spatial distance and coverage progression guide the vehicle into the lane, after which precision angular alignment penalties take effect to ensure exact curb-parallel docking.

\subsection{Unified Terminal and Evaluation Alignment}
To entirely resolve the catastrophic training instabilities where an agent over-optimizes for a "cautious coward" policy—learning to remain completely static outside the parking corridor or spin infinitely in place to safely avoid physical hazards rather than attempting to enter the berth—we explicitly lock the environment's terminal bounds to a unified mathematical formulation. In standard reinforcement learning architectures, a structural mismatch often exists between step-wise dense potentials and the arbitrary thresholds chosen to signal episode success. This disconnect introduces uninformative "dead zones" where an agent, having already achieved an acceptable pose, continues to accumulate minor baseline time penalties and destabilizes its value function network. 

To overcome this optimization bottleneck, we define a continuous global evaluation metric, $S \in [0, 100]$. Unlike typical binary success indicators, this scoring function is formulated to map the entire topology of the final vehicle state. It provides an expressive, scalar performance measure that serves as the invariant objective function for our meta-optimization layer and provides a direct terminal windfall payout. To ensure smooth gradients and avoid step discontinuities that can destabilize optimization algorithms, the metric is structured purely around two operational regimes:
\begin{equation}
S = \begin{cases} 
0.0 & \text{if a collision is detected,} \\
S_{\text{dist}} \times S_{\text{align}} \times S_{\text{lane}} \times 100.0 & \text{otherwise.}
\end{cases}
\end{equation}

\subsubsection{Safety Boundaries and Hard-Failure Conditions}
The primary regime handles absolute safety boundaries. If the vehicle's bounding polygon intersects with any static obstacle (e.g., adjacent parked cars) or breaches the outer road boundaries ($x < 250\,\text{px}$ or $x > 710\,\text{px}$), a hard collision is triggered. Under this condition, the episode terminates immediately ($\text{terminated} = \text{True}$), and the evaluation score drops instantly to $0.0$, ensuring that safety-critical violations are strictly penalized regardless of the vehicle's proximity to the target. The agent receives a flat, negative sparse terminal penalty:
\begin{equation}
R_{\text{terminal}} = -c_{\text{collision}}
\end{equation}
where $c_{\text{collision}}$ is treated as an optimized penalty hyperparameter mapped by the hyperparameter meta-optimization pipeline to effectively balance the maximum step-wise dense gradients explored in prior steps.

\subsubsection{Multiplicative Continuous Alignment}
For all non-colliding terminal states, the score is calculated via a multiplicative assembly of three localized sub-metrics. This continuous scaling maps proximity, alignment, and lane error between fixed zero-point horizons and the ideal target state. As the vehicle approaches a flawless park, the sub-metrics naturally converge to unity, yielding a perfect score of $100.0$:
\begin{enumerate}
    \item \textbf{Proximity Score ($S_{\text{dist}}$):} Measures the Euclidean distance $d$ from the vehicle's center $(x_{\text{car}}, y_{\text{car}})$ to the target center $(x_{\text{target}}, y_{\text{target}})$, normalized against a fixed maximum distance horizon $D_{\max} = 300.0\,\text{px}$:
    \begin{equation}
    S_{\text{dist}} = \max\left(0.0, 1.0 - \frac{d}{D_{\max}}\right)
    \end{equation}
    
    \item \textbf{Alignment Score ($S_{\text{align}}$):} Penalizes the angular deviation $\Delta \theta$ between the vehicle's orientation and the target heading ($\theta_{\text{target}} = -\frac{\pi}{2}$). The error is wrapped between $[-\pi, \pi]$ and normalized against a perpendicular horizon $\Theta_{\max} = \frac{\pi}{2}$:
    \begin{equation}
    S_{\text{align}} = \max\left(0.0, 1.0 - \frac{|\Delta \theta|}{\Theta_{\max}}\right)
    \end{equation}
    
    \item \textbf{Lane Containment Score ($S_{\text{lane}}$):} Measures the lateral displacement of the vehicle's corners relative to the valid parking lane boundaries. To ensure active road clearance, we enforce a strict interior boundary $[X_{\min}, X_{\max}] = [555.0, 705.0]\,\text{px}$. Let $C$ be the set of the vehicle's four bounding corners. The total boundary violation $\epsilon_{\text{lane}}$ is computed as:
    \begin{equation}
    \epsilon_{\text{lane}} = \max_{c \in C} (X_{\min} - c_x, 0.0) + \max_{c \in C} (c_x - X_{\max}, 0.0)
    \end{equation}
    This total pixel violation is normalized against a maximum acceptable error horizon $E_{\max} = 100.0\,\text{px}$:
    \begin{equation}
    S_{\text{lane}} = \max\left(0.0, 1.0 - \frac{\epsilon_{\text{lane}}}{E_{\max}}\right)
    \end{equation}
\end{enumerate}

\subsubsection{Gradient Properties and Optimization Advantages}
The product of these sub-metrics ($S_{\text{dist}} \times S_{\text{align}} \times S_{\text{lane}}$) provides a critical mathematical property for global meta-optimization: \textbf{interdependent directional zeroing}. If an agent achieves exceptional distance proximity but parks completely crooked ($| \Delta \theta | \ge \frac{\pi}{2}$) or leaves the vehicle body sticking out into active road traffic ($\epsilon_{\text{lane}} \ge 100.0\,\text{px}$), the corresponding sub-metric drops to $0.0$, collapsing the entire global score to zero.

By removing manual piece-wise success jumps, we eliminate artificial cliffs and gradient spikes within the search space. Within the valid operating envelope, this pure formulation provides smooth, continuous partial derivatives that supply clear directional feedback, guiding sequence-based or surrogate-based meta-optimization algorithms efficiently toward the global maximum.

To close the loop with the reinforcement learning agent, an early success condition ($\text{terminated} = \text{True}$) is triggered automatically if and only if the absolute evaluation score exceeds the optimized quality cutoff threshold managed dynamically by the hyperparameter pipeline:
\begin{equation}
S > \mathcal{S}_{\text{threshold}}
\end{equation}
where $\mathcal{S}_{\text{threshold}}$ represents the performance gate evaluated during environment execution. Upon crossing this threshold, the step rewards freeze, and the agent instantly bankrolls a positive sparse terminal windfall:
\begin{equation}
R_{\text{terminal}} = S \cdot w_{\text{success}}
\end{equation}
where $w_{\text{success}} = 1.0$. By binding step-wise potential convergence, early success criteria, and meta-optimization targets to the exact same continuous mathematical function $S$, the algorithm's value function estimator achieves total alignment, accelerating policy updates.

\begin{figure}[H]
    \centering
    \begin{subfigure}[b]{0.19\textwidth}
        \centering
        \includegraphics[width=\textwidth]{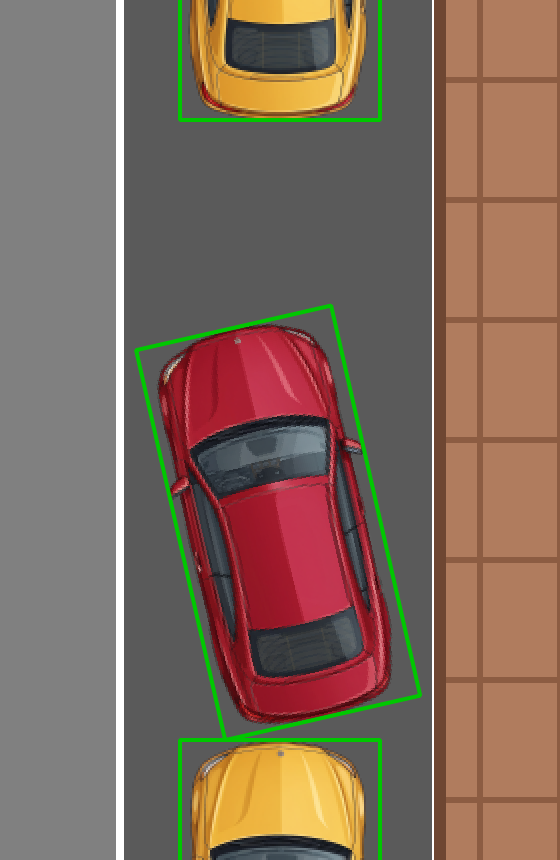}
        \caption{$S = 0.0$ (Collision)}
        \label{fig:score1}
    \end{subfigure}
    \hfill
    \begin{subfigure}[b]{0.19\textwidth}
        \centering
        \includegraphics[width=\textwidth]{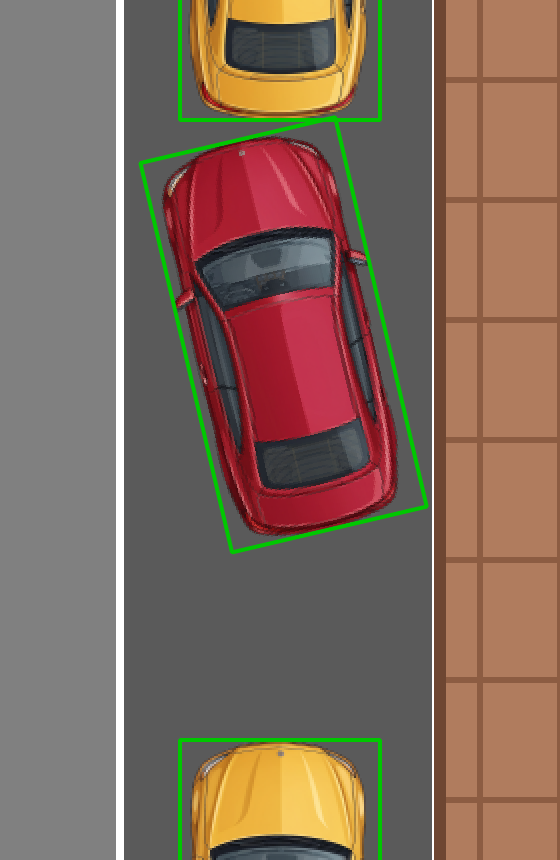}
        \caption{$S = 0.0$ (Collision)}
        \label{fig:score2}
    \end{subfigure}
    \hfill
    \begin{subfigure}[b]{0.19\textwidth}
        \centering
        \includegraphics[width=\textwidth]{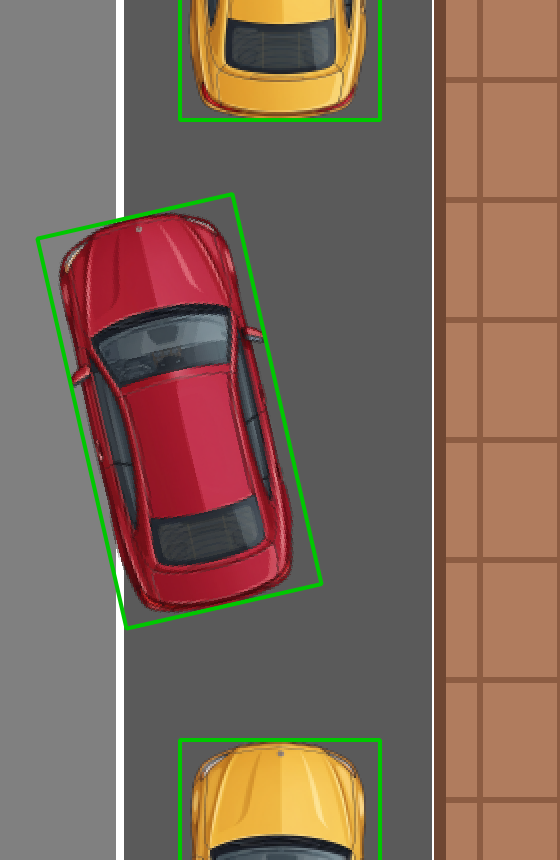}
        \caption{$S = 38.2$}
        \label{fig:score3}
    \end{subfigure}
    \hfill
    \begin{subfigure}[b]{0.19\textwidth}
        \centering
        \includegraphics[width=\textwidth]{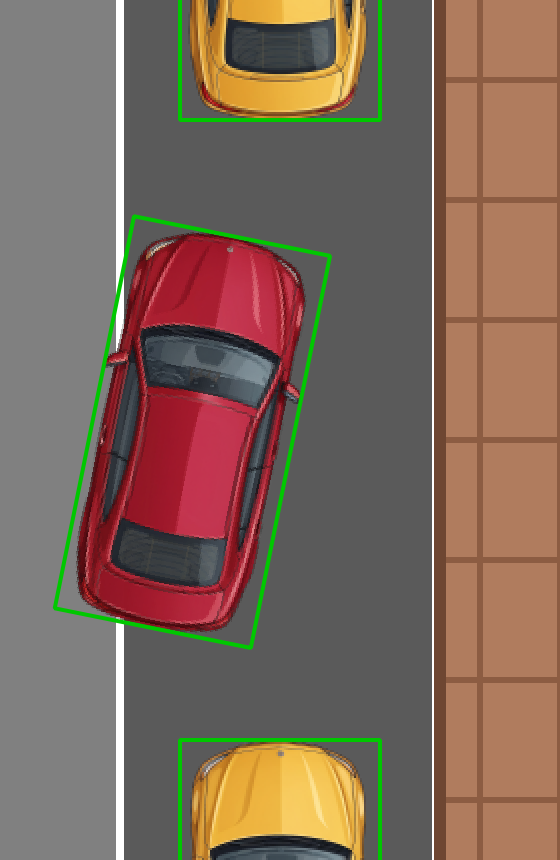}
        \caption{$S = 46.5$}
        \label{fig:score4}
    \end{subfigure}
    \hfill
    \begin{subfigure}[b]{0.19\textwidth}
        \centering
        \includegraphics[width=\textwidth]{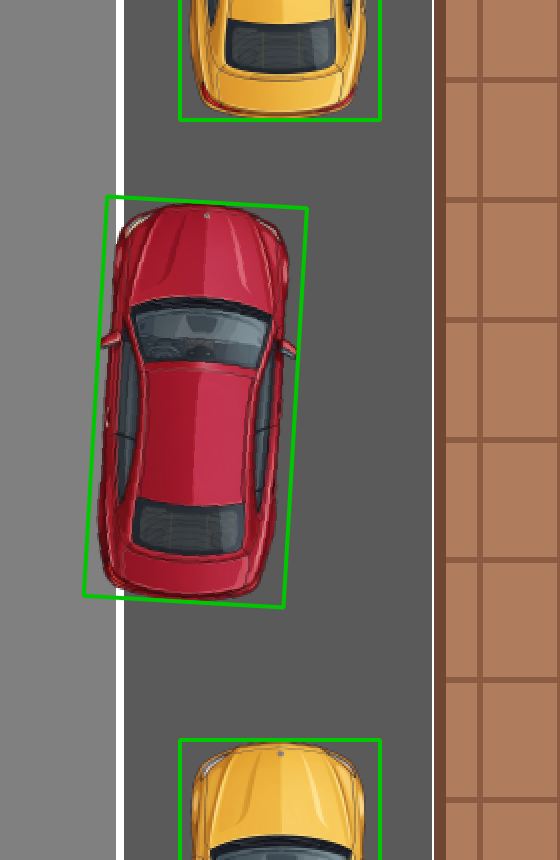}
        \caption{$S = 63.0$}
        \label{fig:score5}
    \end{subfigure}
    
    \vspace{1em} 
    
    \begin{subfigure}[b]{0.19\textwidth}
        \centering
        \includegraphics[width=\textwidth]{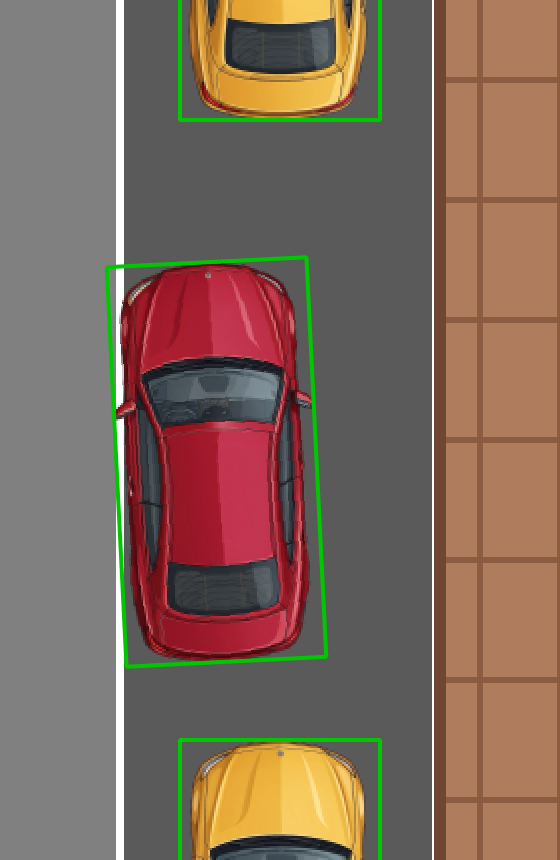}
        \caption{$S = 75.4$}
        \label{fig:score6}
    \end{subfigure}
    \hfill
    \begin{subfigure}[b]{0.19\textwidth}
        \centering
        \includegraphics[width=\textwidth]{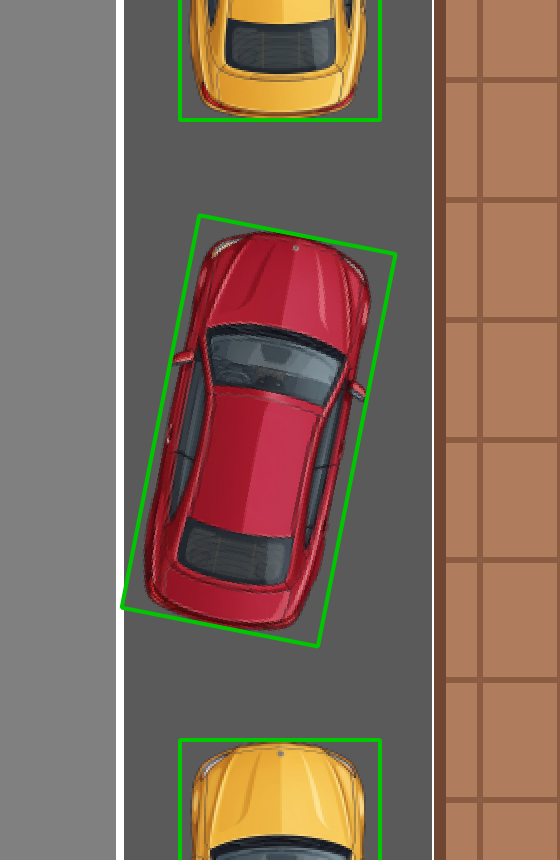}
        \caption{$S = 80.9$}
        \label{fig:score7}
    \end{subfigure}
    \hfill
    \begin{subfigure}[b]{0.19\textwidth}
        \centering
        \includegraphics[width=\textwidth]{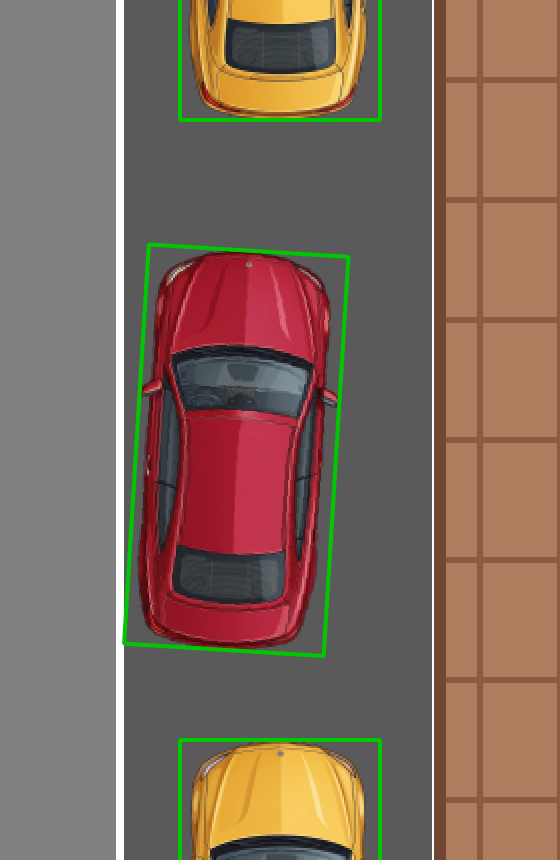}
        \caption{$S = 85.7$}
        \label{fig:score8}
    \end{subfigure}
    \hfill
    \begin{subfigure}[b]{0.19\textwidth}
        \centering
        \includegraphics[width=\textwidth]{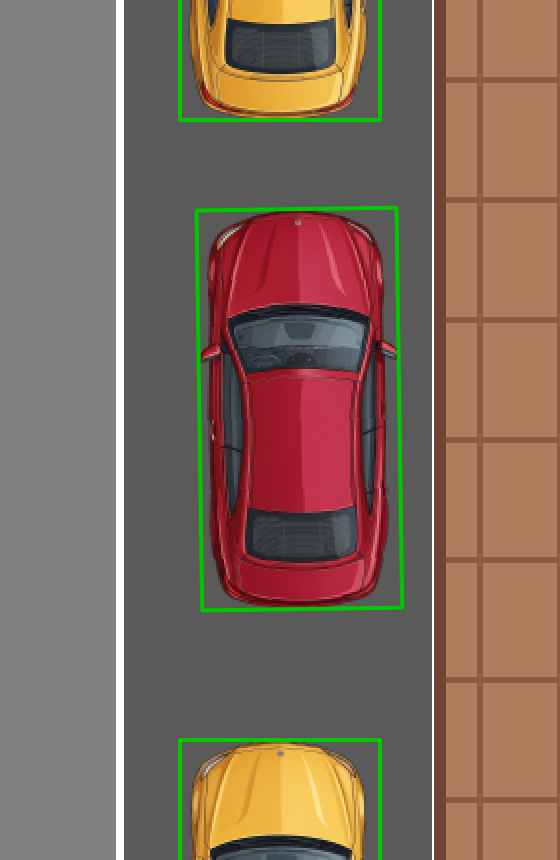}
        \caption{$S = 94.4$}
        \label{fig:score9}
    \end{subfigure}
    \hfill
    \begin{subfigure}[b]{0.19\textwidth}
        \centering
        \includegraphics[width=\textwidth]{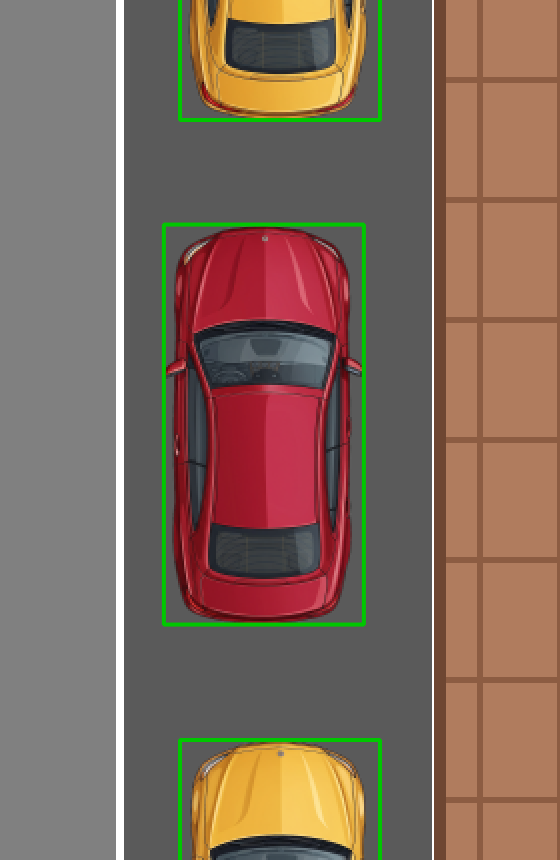}
        \caption{$S = 97.2$}
        \label{fig:score10}
    \end{subfigure}
    
    \caption{A gallery of ten representative final vehicle poses evaluated by the continuous scoring function. The images trace the graduation from hard safety failures ($S=0.0$) through partial geometric alignments up to an ideal, precision parallel park ($S=100.0$).}
    \label{fig:score_gallery}
\end{figure}

To demonstrate the practical behavior of the multiplicative grading mechanics, Figure~\ref{fig:score_gallery} illustrates ten distinct final vehicle states generated during evaluation. The gallery highlights how safety violations immediately drop the objective function to zero, while varying degrees of proximity, alignment, and lane containment map smoothly across the continuous $[0, 100]$ spectrum.

\section{Reinforcement Learning Training Methodology}
With the Markov Decision Process (MDP) and the parameterized dynamic reward landscape formally defined, this section outlines the core training architecture. The implementation integrates value-based deep reinforcement learning with an explicit, geometry-driven stochastic spawn curriculum to systematically optimize the policy over diverse initializations.

\subsection{Deep Q-Network (DQN) Core Architecture}
To synthesize optimal control policies within the 9-way discrete action space $\mathcal{A}$, the agent leverages a Deep Q-Network (DQN) implementation \cite{mnih2015human} built on the \texttt{stable-baselines3} framework \cite{stable-baselines3}. The network approximates the optimal action-value function $Q^*(s, a)$ via a Multi-Layer Perceptron (MLP) consisting of two hidden layers with $256$ units each, mapping the 21-dimensional normalized observation vector directly to action Q-values.

Temporal difference learning updates are stabilized using a uniform experience replay buffer $\mathcal{B}$ with a capacity of $200,000$ transitions and a decoupled target network $Q(s, a; \theta^{-})$ updated hard-periodically every $10,000$ steps ($\tau = 1.0$) with a discount factor $\gamma = 0.99$. Exploration is governed by an adaptive $\epsilon$-greedy mechanism, where the probability of selecting a random action decays linearly over an optimized exploration fraction of the $10^7$ step training budget down to a clamped lower bound $\epsilon_{\text{final}}$. Both the learning rate $\alpha$ and the exploration decay schedule are managed directly as design variables by the outer meta-optimization framework.

\subsection{Stochastic Spatial Spawn Curriculum}
A critical challenge in training non-holonomic agents for parallel or reverse docking maneuvers is the fragility of random exploration under severe geometric constraints. If an agent is initialized exclusively on the far roadway, the probability of executing the highly precise sequence of steering corrections required to enter a narrow, obstacle-bounded parking berth by chance approaches zero. To alleviate this exploratory bottleneck, the environment implements a multi-stage, geometry-driven spatial spawn curriculum activated exclusively during the training phase via the \texttt{enable\_training\_spawns} control state.

At the initialization of each training episode, the environment samples a uniform random variable $\mu \sim U(0,1)$ via a stochastic filter to dynamically shift the agent's spatial initial conditions across three distinct tactical operational regimes:
\begin{enumerate}
    \item \textbf{Parallel Driving Lane Spawning (50\% Probability Baseline, $\mu \ge 0.5$):} The vehicle spawns natively in the main traffic lane centered at the starting anchors ($x_{\text{start}}, y_{\text{start}}$) with a continuous uniform coordinate variation of $\pm 25.0$\,px along the longitudinal axis ($x \in [x_{\text{start}}-25, x_{\text{start}}+25]$). The orientation is bound strictly to the horizontal road axis ($\theta_{\text{target}}$). This layout demands a complete multi-step maneuvering sequence from the primary roadway position.

    \item \textbf{Curb-Border Entry Manipulation (40\% Probability Shift, $\mu < 0.4$):} The environment overrides the baseline road initialization via a dedicated border entry sampler. The vehicle is positioned on an intermediate trajectory near the parking slot boundary, forcing the policy to directly master terminal reverse entry sweeps and spatial arc alignment.

    \item \textbf{Micro-Berth Docking Adjustments (10\% Probability Shift, $0.4 \le \mu < 0.5$):} The vehicle is placed directly within the interior bounds of the destination parking berth via an internal spawn module. This configuration isolates fine-grained micro-alignment corrections, requiring the agent to learn precise terminal straightening and stationary holding actions without executing the preceding roadway navigation phase.
\end{enumerate}

\begin{figure}[htbp]
    \centering
    \begin{subfigure}[b]{0.35\linewidth}
        \centering
        \includegraphics[width=\linewidth]{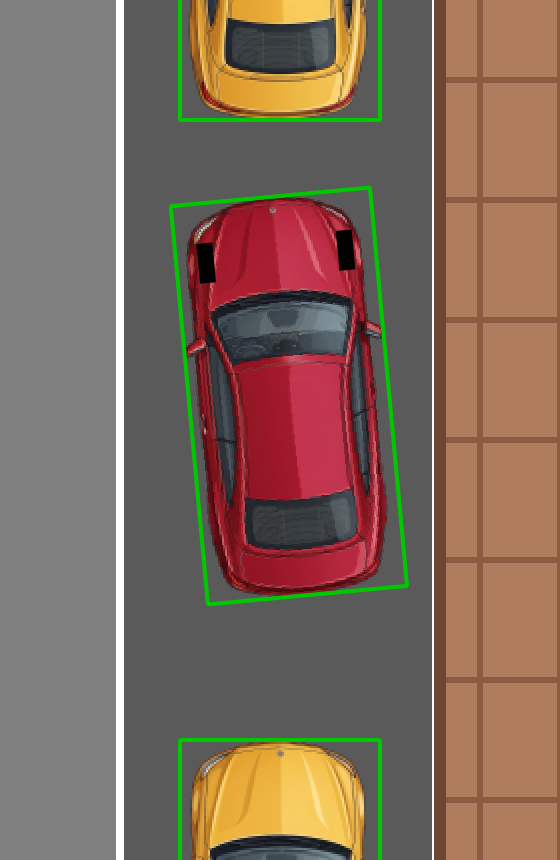}
        \caption{Micro-adjustments}
        \label{fig:stage1_spawn}
    \end{subfigure}%
    \hspace{1em}%
    \begin{subfigure}[b]{0.35\linewidth}
        \centering
        \includegraphics[width=\linewidth]{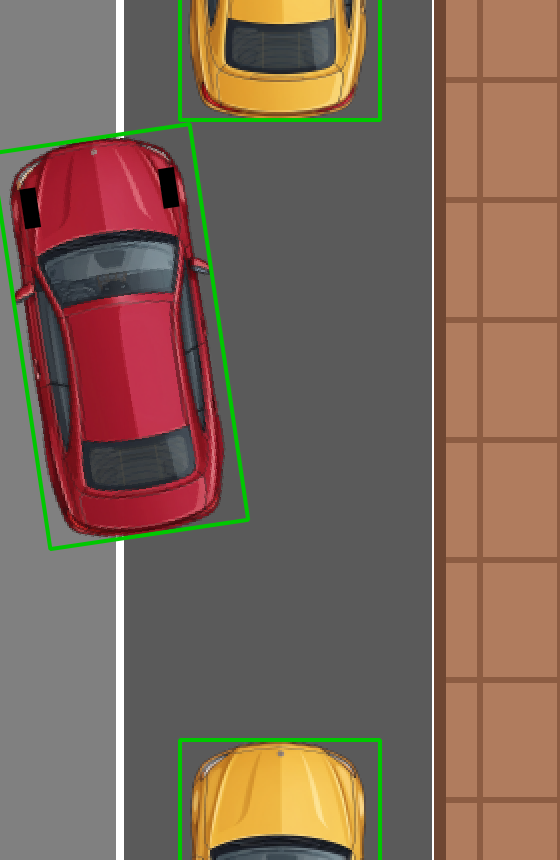}
        \caption{On the parking lane border}
        \label{fig:stage2_spawn}
    \end{subfigure}
    \caption{Stochastic multi-stage spawn curriculum: (a) Micro-berth docking adjustments, and (b) Reverse entry from the parking lane border.}
    \label{fig:curriculum_stages}
\end{figure}

To preserve environment safety and prevent the injection of invalid initial training frames, intermediate positioning and lane collision layouts are validated through automated safety filters. When training spawns are deactivated, the environment bypasses the stochastic filter completely, locking all initializations to the full-difficulty operational driving lane baseline to guarantee strict, unbiased evaluation metrics.

\subsection{Software Stack Dependencies}
The implementation is built upon a Python open-source framework, relying on a modular pipeline:
\begin{itemize}
    \item \textbf{Gymnasium Environment Wrapper:} The environment is formalized using the Farama Foundation's \texttt{gymnasium} core package, implementing standardized \texttt{reset()} and \texttt{step()} methods, alongside standardized \texttt{spaces.Box} observation vectors and \texttt{spaces.Discrete} joint action matrices.
    \item \textbf{Stable-Baselines3 (SB3):} High-performance, tensorboard-integrated algorithm implementations are driven by the \texttt{stable-baselines3} framework, wrapping the execution logic into optimized, vector-ready wrappers.
\end{itemize}

\subsection{Decoupled Generalization Cross-Validation}
To guarantee that the hyperparameter vector $\Phi$ optimized by the outer-loop engine yields genuine navigation capabilities rather than an artifact of over-fitted training initializations, the policy metrics are strictly decoupled from the final evaluation routine. While the inner-loop training uses the 3-stage spawn curriculum alongside continuous dense potential gradients over $10,000,000$ simulation steps, the evaluation phase completely disables these components. 

Upon completion of the training budget, the policy weights are frozen, and the agent is then evaluated across $1,000$ independent games played from randomized spawn variations along the unconstrained roadway. Each trial is scored by the continuous invariant multi-criteria quality score $S \in [0, 100]$. The empirical mean performance of these $1,000$ trials is the final objective score used by the \texttt{rodopt+} optimizer to navigate the high-level parameter space.

\section{Hyperparameter Meta-Optimization Framework}
The performance of model-free reinforcement learning algorithms in highly constrained non-holonomic environments is notoriously sensitive to hyperparameter selection. In our autonomous parking vehicle scenario, this vulnerability is compounded by the structural interdependence between standard algorithmic parameters (e.g., learning mechanics, exploration rules) and environment-specific reward-shaping parameters (e.g., dynamic weight crossover thresholds, sparse boundary penalties). Manual heuristic tuning of such coupled systems invariably yields suboptimal local minima, such as the twins of premature immobilization and erratic orientation oscillations.

To systematically address this limitation, we formalize the calibration phase as a black-box meta-optimization problem where the previously introduced continuous trajectory quality metric, $S \in [0, 100]$, serves directly as the objective function. Let $\Phi \in \mathbb{R}^{D}$ denote a $D$-dimensional design vector containing the hyperparameter configurations ($D = 14$). Evaluating a candidate vector $\Phi$ requires executing an entire inner-loop reinforcement learning training cycle spanning a rigorous horizon of $10,000,000$ simulation time steps. 

To ensure that the optimized parameters yield a policy that generalizes robustly across the state space rather than overfitting to a static initialization, the trained agent is subjected to a comprehensive evaluation phase consisting of $1,000$ distinct games. In each evaluation game, the vehicle's initial roadway spawn coordinates are stochastically perturbed. The global black-box objective function $f(\Phi)$ is defined as the empirical mean evaluation score across these $N = 1,000$ randomized trials. The meta-optimization problem is formulated as:
\begin{align}
    \max_{\Phi} \quad & f(\Phi) = \frac{1}{N} \sum_{g=1}^{N} S_g(\Phi) \\
    \text{s.t.} \quad & \Phi_{\min}^{(i)} \le \Phi^{(i)} \le \Phi_{\max}^{(i)}, \quad \forall i \in \{1, \dots, D\}
\end{align}

\subsection{Algorithmic Pipeline via \texttt{rodopt+}}
Evaluating the objective function $f(\Phi)$ is computationally intensive, as it embeds a full $10^7$-step training sequence within each function evaluation. To solve this problem efficiently within a tight global computational budget of $500$ total function evaluations, we utilize the proprietary optimization engine \texttt{rodopt+} (\url{www.rodopt.com}). The workflow leverages an outer-loop architecture split into two structural regimes: space-filling initialization and surrogate-driven Bayesian optimization.

\subsubsection{Space-Filling Initialization}
To eliminate initialization bias and construct a robust baseline coverage layout across the high-dimensional bounded search space, \texttt{rodopt+} leverages a Latin Hypercube Sampling (LHS) method. The 14-dimensional parameter space is partitioned into orthogonal sub-intervals, ensuring that each parameter dimension is sampled uniformly without redundant coordinate projections. This initial sampling phase populates the exploratory database with well-distributed data points before the active optimization surrogate loop takes over.

\subsubsection{Surrogate-Based Bayesian Optimization}
Following the initialization phase, \texttt{rodopt+} transitions into a sequential Bayesian optimization (BO) regime designed specifically for expensive black-box objective functions. BO serves as an ideal paradigm for reinforcement learning hyperparameter calibration due to its exceptional sample efficiency, constructing a probabilistic statistical model of the objective landscape rather than relying on dense random sampling or finite-difference gradient calculations. Furthermore, the BO framework in \texttt{rodopt+} natively scales to modern parallel computing architectures because candidate evaluations can be executed in a fully asynchronous manner, ensuring continuous cluster utilization without idle worker blocking. The underlying architecture is also highly modular, making it easily extensible to advanced optimization paradigms such as explicit constraint handling, robust design under domain shifts, and multi-objective Pareto studies.

In our case study, the algorithm constructs a probabilistic Gaussian Process (GP) surrogate model $\mathcal{G}(\Phi)$ to approximate the unknown fitness landscape of the mean evaluation score $f(\Phi)$ based on design-performance pairs collected in prior iterations. At each global optimization increment, an acquisition function—such as Expected Improvement (EI)—is maximized over the GP model to resolve the trade-off between exploitation (sampling regions where the predicted mean trajectory quality score is high) and exploration (sampling regions with high surrogate uncertainty).

\subsection{Optimization Parameters and Infrastructure}
The 14 design variables passed from the outer optimization layer are classified into two decoupled tracking domains, mapped out between explicit continuous operational boundaries:
\begin{enumerate}
    \item \textbf{Algorithmic Execution Parameters (3 variables):} The neural network learning rate is constrained within a real-valued spectrum $\alpha \in [10^{-5}, 10^{-3}]$. Policy exploration is parameterized by an exploration fraction bounded within $[0.2, 0.5]$ alongside a clamped final terminal exploration epsilon range $\epsilon_{\text{final}} \in [0.01, 0.05]$.
    \item \textbf{Reward Architecture and Terminal Parameters (11 variables):} Environment boundary thresholds, dynamic potential field weights, action regularization penalties, and gating conditions are managed dynamically. This includes the maximum collision penalty magnitude $c_{\text{collision}} \in [0.0, 50.0]$, the spatial distance penalty weight $w_{\text{dist}} \in [0.0, 2.0]$, the angular alignment penalty weight $w_{\text{align}} \in [0.0, 3.0]$, the baseline time penalty magnitude $c_{\text{baseline}} \in [0.001, 0.1]$, and the drive-direction switch penalty $c_{\text{switch}} \in [0.0, 5.0]$. Corridor insertion shaping is bounded by the standing existence bonus weight $w_{\text{standing}} \in [0.0, 1.0]$ and directional progress weight $w_{\text{delta}} \in [0.0, 5.0]$, while coverage activation triggers span $C_{\text{low}} \in [0.0, 0.3]$ and $C_{\text{high}} \in [0.8, 1.0]$. Finally, early episode completion is governed by the success gate score threshold $\mathcal{S}_{\text{threshold}} \in [94.0, 99.9]$ and the terminal success windfall payout scale $w_{\text{success}} \in [50.0, 300.0]$.
\end{enumerate}

Data communication between the optimization engine and the model-free training process is managed through a structured, decoupled data transfer interface. At the start of a design iteration, the optimizer generates a unique design candidate vector $\Phi_k$ and exports the configuration parameters. 

The evaluation framework then invokes the reinforcement learning execution module. The pipeline registers the candidate hyperparameter vector, instantiates the simulation environment variables, and executes the $10,000,000$ step inner-loop training phase. Upon completion, the trained policy is automatically evaluated across 1,000 validation trials initialized from randomized roadway spawn configurations. Trajectories are scored against the invariant trajectory quality metric, and the empirical mean score is returned to the optimization engine. The surrogate model parses this objective output, updates its Gaussian Process history, and dynamically computes the subsequent design vector to systematically maximize overall docking performance.

\section{Results and Discussion}

\subsection{Quantitative Performance and Convergence}
To evaluate the impact of bi-level hyperparameter optimization on policy learning, the Deep Q-Network (DQN) agent was benchmarked across the parameter space using a structured Design of Experiments (DoE) sampling framework. Each candidate reward configuration was trained over $10^7$ simulation steps and subsequently evaluated across $1,000$ randomized validation trials initialized from main roadway spawn poses.

Across the overall DoE sampling space—comprising uncalibrated initializations, default scalar weights, and random parameter configurations—the policy exhibited severe convergence bottlenecks, yielding an empirical average mean score of \textbf{26.7 / 100.0}. This low baseline performance reflects the exploratory fragility and localized gradient traps inherent to non-holonomic parallel docking under naive, unoptimized reward scalar combinations.

In contrast, when trained under the optimal reward structure identified by our optimization framework, policy performance increased dramatically, achieving a mean evaluation score of \textbf{95.6 / 100.0}. This result confirms that systematic potential field shaping and action regularization enable the model-free agent to reliably negotiate narrow geometric corridors, maintaining precise parallel curb alignment and controlled deceleration without colliding with adjacent parked vehicles or pavement borders.

\subsection{Necessity of Reward Parameter Optimization}
A central finding of this investigation is the absolute necessity of systematic hyperparameter optimization for policy convergence in constrained control tasks. In uncalibrated baseline configurations—corresponding to the low sampling baseline mean score of $26.7$—imbalanced scalar trade-offs between continuous distance surrogates, alignment potential fields, directional progress rewards, and collision penalties caused the policy to degenerate into characteristic sub-optimal behaviors:

\begin{enumerate}
    \item \textbf{Jittery Immobilization in the Parking Lane:} When angular alignment penalties ($w_{\text{align}}$) or localized boundary costs are improperly scaled against coverage progress gains, the vehicle enters the parking lane but experiences policy paralysis. To avoid incurring orientation misalignment penalties while maximizing step-wise existence rewards, the agent executes high-frequency, oscillating micro-steers or tiny forward-reverse toggles, remaining jittery and stuck in the spot without straightening or completing the maneuver.
    \item \textbf{Over-Conservative Hazard Avoidance:} When the collision penalty ($c_{\text{collision}}$) dominates the terminal success payout ($w_{\text{success}}$) and potential guidance gains, penalty dominance occurs. During early exploration, the agent learns that entering the narrow berth between parked obstacles incurs a high risk of catastrophic negative feedback. Consequently, the policy optimizes purely for hazard avoidance, remaining idle on the open roadway or drifting away from the parking berth to guarantee safety at the expense of task execution.
\end{enumerate}

Systematically tuning these parameter weights and decay curves via bi-level optimization resolves the trade-off between safety constraints and goal progression, generating a smooth optimization landscape that guides the agent toward full task completion.

\subsection{Training Dynamics and Convergence Behavior}
\label{subsec:training_curves}

\begin{figure}[htbp]
    \centering
    \includegraphics[width=0.85\linewidth]{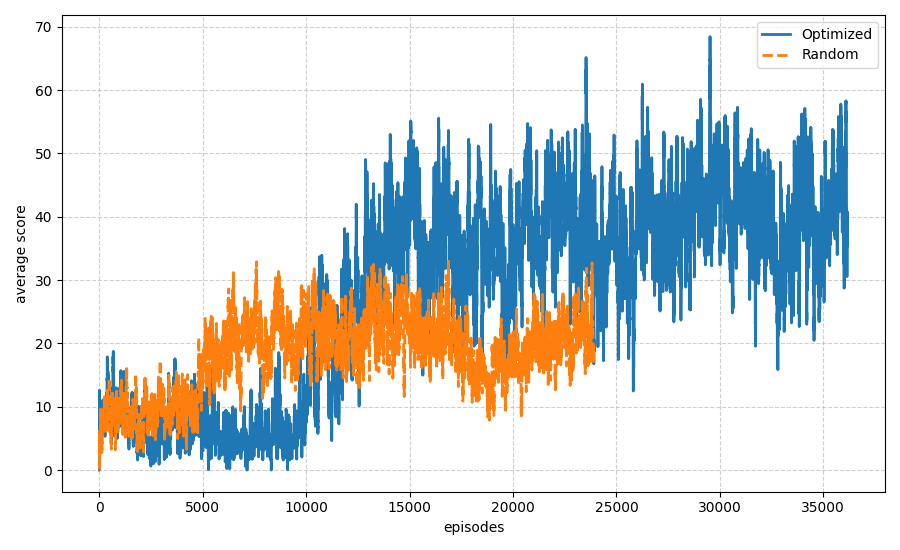} 
    \caption{Moving average episode score during training ($10^7$ simulation timesteps). The blue curve denotes the policy trained under the optimized reward field parameters, whereas the orange curve represents an unoptimized baseline configuration ($\bar{S} \approx 34.4$).}
    \label{fig:training_moving_avg}
\end{figure}

To evaluate the impact of optimal reward shaping on policy convergence and sample efficiency, \Cref{fig:training_moving_avg} contrasts the moving average episode score during training between the optimized reward parameters and a representative unoptimized (random) baseline configuration. The baseline configuration mirrors the mean reward performance ($\bar{S} \approx 34.4$) recorded during the initial Design of Experiments (DoE) sampling phase.

As shown in \Cref{fig:training_moving_avg}, the optimized policy (blue curve) consistently outperforms the unoptimized baseline (orange curve) throughout the entire training horizon. Notably, two structural phenomena characterize these trajectories:

\begin{enumerate}
    \item \textbf{Episode Throughput and Truncation Rates:} Within the fixed simulation horizon of $10^7$ environment steps, the optimized configuration completes a significantly higher total number of episodes compared to the baseline. Under the unoptimized reward structure, the agent repeatedly becomes trapped in non-holonomic local minima or execution stalls, frequently reaching the maximum horizon limit ($T_{\text{max}} = 500$ steps) and triggering episode truncation. In contrast, the shaped dynamic reward incentivizes early task completion, allowing the agent to park efficiently and proceed to subsequent training episodes at an accelerated rate.
    
    \item \textbf{Impact of Exploratory Noise during Training:} Although the final evaluation score of the optimized policy approaches near-perfect performance ($95.6$), its training moving average stabilizes at a lower equilibrium (approximately $50.0$). This discrepancy is a direct artifact of the $\epsilon$-greedy exploration mechanism active during training. Stochastic action sampling introduces deliberate off-policy perturbations and random collisions necessary for state-space exploration. Once training completes and exploratory noise is disabled ($\epsilon = 0$) during the final $1,000$ deterministic evaluation episodes, the policy demonstrates its true optimal control behavior without exploratory penalties.
\end{enumerate}

\subsection{Visual Trajectory Demonstration}
A supplementary video demonstration illustrating the optimized Deep Q-Network policy executing the complete autonomous parallel parking maneuver is available online.\footnote{Video demonstration: \url{https://www.youtube.com/watch?v=TJeT37g6564}} The visual trajectory confirms smooth entry arc generation, continuous speed regulation, and precise terminal alignment within the target berth. The simulation scripts and instructions to reproduce this trajectory are available in the repository cited in the Code Availability section.

\section{Conclusion}
This paper presented a parameterized reward shaping framework paired with an aligned termination mechanism to eliminate fundamental exploration failures in non-holonomic reinforcement learning tasks. Through joint hyperparameter optimization of reward constants and deep learning parameters using Bayesian optimization, our Deep Q-Network agent successfully improved the mean evaluation score in a complex autonomous parallel parking environment. These results confirm that systematic co-optimization is a mandatory requirement for stable policy convergence in tightly constrained control domains.

\section*{Acknowledgments}
The authors gratefully acknowledge the use of generative AI tools during the development of this work, specifically Gemini (Google) and GitHub Copilot (Microsoft). GitHub Copilot was utilized for code generation, auto-completion, and refactoring during software implementation. Gemini was used as a research collaborator to assist with environment class structuring, physics simulation logic, copy-editing, refining \LaTeX{} formatting, and improving prose clarity. All core conceptual frameworks, mathematical formulations, algorithmic implementations, and experimental results were conceived, executed, and verified by the authors, who take full responsibility for the final content of this paper.

\section*{Code and Media Availability}
The source code for the simulation environment, reinforcement learning training pipeline, and evaluation scripts used in this study are publicly available on GitHub at \url{https://github.com/eoezkaya/ParkingGame.git}. The repository \texttt{README} provides step-by-step execution details to reproduce the training procedures, policy evaluation, and visual trajectory demonstrations.

\bibliography{sample}

@article{song2020data,
  title={Data efficient reinforcement learning for integrated lateral planning and control in automated parking system},
  author={Song, Shuo and Chen, Hong and Sun, Haobin and Liu, Meilan},
  journal={Sensors},
  volume={20},
  number={24},
  pages={7297},
  year={2020},
  publisher={MDPI}
}

@article{tian2026hierarchical,
  title={A Hierarchical NMPC and TD3-Based Framework for Seamless Cruise-to-Park Automated Valet Parking},
  author={Tian, D.},
  journal={Sensors},
  volume={26},
  number={11},
  pages={3409},
  year={2026},
  publisher={MDPI}
}

@article{berta2024development,
  title={Development of Deep-Learning-Based Autonomous Agents for Low-Speed Maneuvering in Unity},
  author={Berta, Riccardo and Lazzaroni, Luca and Capello, Alessio and Cossu, Marianna and Forneris, Luca and Pighetti, Alessandro and Bellotti, Francesco},
  journal={Journal of Intelligent and Connected Vehicles},
  volume={7},
  number={3},
  pages={229--244},
  year={2024},
  publisher={Tsinghua University Press}
}

@article{huang2020closer,
  title={A Closer Look at Invalid Action Masking in Policy Gradient Algorithms},
  author={Huang, Costa and Ontan{\'o}n, Santiago},
  journal={arXiv preprint arXiv:2006.14171},
  year={2020}
}

@book{sutton2018reinforcement,
  title={Reinforcement learning: An introduction},
  author={Sutton, Richard S and Barto, Andrew G},
  year={2018},
  publisher={MIT press}
}

@inproceedings{abel2015goal,
  title={Goal-Based Action Priors},
  author={Abel, David and Hershkowitz, D Ellis and Koppula, Windee and Brawner, Gabriel and MacGlashan, James and Littman, Michael L},
  booktitle={Proceedings of the International Conference on Automated Planning and Scheduling},
  volume={25},
  pages={306--314},
  year={2015}
}

@article{mnih2015human,
  title={Human-level control through deep reinforcement learning},
  author={Mnih, Volodymyr and Kavukcuoglu, Koray and Silver, David and Rusu, Andrei A and Veness, Joel and Bellemare, Marc G and Graves, Alex and Riedmiller, Martin and Fidjeland, Andreas K and Ostrovski, Georg and others},
  journal={Nature},
  volume={518},
  number={7540},
  pages={529--533},
  year={2015},
  publisher={Nature Publishing Group}
}

@article{stable-baselines3,
  author  = {Antonin Raffin and Ashley Hill and Adam Gleave and Anssi Kanervisto and Maximilian Ernestus and Noah Dormann},
  title   = {Stable-Baselines3: Reliable Reinforcement Learning Implementations in Python},
  journal = {Journal of Machine Learning Research},
  year    = {2021},
  volume  = {22},
  number  = {268},
  pages   = {1-8}
}

\end{document}